\documentclass{article}
\usepackage{amsmath,epsfig}
\usepackage[preprint]{spconfa4}

\usepackage{microtype}
\usepackage{subfigure}
\usepackage{booktabs} 
\usepackage{multicol}
\usepackage{multirow}
\usepackage{diagbox}
\usepackage{enumitem}

\let\OLDthebibliography\thebibliography
\renewcommand\thebibliography[1]{
  \OLDthebibliography{#1}
  \setlength{\parskip}{0pt}
  \setlength{\itemsep}{0pt plus 0.3ex}
}

\begin{document}\sloppy

\def\x{{\mathbf x}}
\def\L{{\cal L}}

\title{Rethinking Hard-Parameter Sharing in Multi-Domain Learning}
%
\name{Lijun Zhang$^{\ast}$, Qizheng Yang$^{\dagger}$, Xiao Liu$^{\ddagger}$, Hui Guan$^{\diamond}$}
\address{University of Massachusetts Amherst, Amherst, MA, USA \\
$^{\ast}$ lijunzhang@cs.umass.edu; 
$^{\dagger}$ yangqizheng@cs.umass.edu; \\
$^{\ddagger}$ xiaoliu1990@cs.umass.edu;
$^{\diamond}$ huiguan@cs.umass.edu.}

\maketitle

\begin{abstract}
Hard parameter sharing in multi-domain learning (MDL) allows domains to share some of the model parameters to reduce storage cost while improving prediction accuracy. One common sharing practice is to share the bottom layers of a deep neural network among domains while using separate top layers for each domain. In this work, we revisit this common practice via an empirical study on image classification tasks from a diverse set of visual domains and make two surprising observations. (1) Using separate bottom-layer parameters could achieve significantly better performance than the common practice and this phenomenon holds with different experimental settings. (2) A multi-domain model with a small proportion of domain-specific parameters from bottom layers can achieve competitive performance with independent models trained on each domain separately. Our observations suggest that people adopt the new strategy of using separate bottom-layer parameters as a stronger baseline for model design in MDL.
\end{abstract}
\begin{keywords}
Multi-domain Learning, Hard-parameter Sharing, Empirical Study
\end{keywords}

\section{Introduction}
Recent years have witnessed the rapid development of Deep Neural Network (DNN) and their superior performance in many areas of artificial intelligence (AI) and vision tasks. AI-powered applications thus increasingly adopt DNNs for solving tasks in single or multiple data steams, leading to more than one DNN model running simultaneously on resource-constrained embedded devices.  Although the recent progress on efficient model design and model compression has made it easier to deploy a single model on device, supporting many models on device is still challenging due to the linearly increased bandwidth, energy, and storage costs.

An effective approach to address the problem is multi-domain learning (MDL), where several different domains are learned jointly. 
Compared to learning each domain separately, MDL allows some parameter sharing across domains to take advantage of the domains' similarities for improved task performance and reduced storage cost. 
MDL is closely related to multi-task learning (MTL), where a set of tasks are learned jointly. Typically, MTL refers to learning different downstream tasks (e.g., depth estimation and semantic segmentation) together while MDL aims to learn on multiple datasets (e.g., data collected from multiple sources with differing statistics) addressing tasks corresponding to each dataset simultaneously. Despite the subtle differences, both MDL and MTL face an open question on how to share parameters across domains or tasks. In this paper, we use the term ``domain'' and ``task'' interchangeably as each domain is associated with one task.

\begin{figure}[htb]
\begin{center}
\centerline{\includegraphics[scale=0.45]{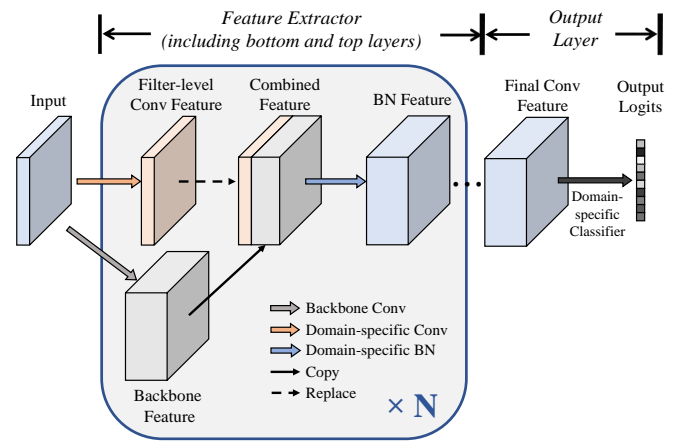}}
\caption{The forward propagation of a multi-domain model given a domain's input. The model consists of domain-specific convolution filters, BN layers, and a classifier. When a set of filters in a convolutional layer is domain-specific, the activation maps from these filters (called filter-level conv feature) will replace the corresponding activation maps from the backbone architecture (called backbone feature). The combined feature will be feed into the next layer.}
\label{fig:methodology}
\vspace{-0.1in}
\end{center}
\end{figure}

Inspired by the effectiveness of the hard parameter sharing strategy in MTL for improving task performance, researchers believe the same strategy is also a promising solution for MDL. Hard parameter sharing is generally applied by sharing the bottom layers among all tasks, while keeping several top layers and an output layer task-specific~\cite{ruder2017overview}. It is commonly used in designing multi-task DNN models in the literature \cite{long2017learning, ruder2019latent} and gains popularity in MDL \cite{sarwar2019incremental, he2020multi}.  

However, it is unclear whether the effectiveness of sharing parameters in bottom layers in MTL can transfer to MDL or not. As shown in Figure~\ref{fig:methodology}, a DNN model usually relies on a stack of layers to transform inputs to features and then an output layer to produce predictions based on the features. Researchers~\cite{mahendran2016visualizing, qin2018convolutional} believe that the first several layers (bottom layers) serve as low-level feature extractors such as edge detectors and corner detectors, which should be able to be shared across multiple tasks. On the contrary, the top layers are more sensitive to the input data (i.e., the training task), implying that different tasks should possess distinct top layers to generate diverse high-level features. 
This belief is intuitively reasonable but lacks solid experimental verification for tasks in diverse domains. Whether the conventional hard parameter sharing strategy could lead to the satisfying performance in MDL remains an open question.

In this work, we show that the above sharing strategy is not suitable in MDL through an empirical study on fine-grained image classification tasks and a popular multi-domain learning benchmark Decathlon \cite{rebuffi2017learning}. Specifically, the common hard parameter sharing strategy, which shares low-level (bottom-layer) parameters while keeping high-level (top-layer) ones domain-specific, is compared to its counterpart, which uses separate bottom-layer parameters for each domain and shares top-layer ones. For the description purpose, we refer to the common hard parameter sharing strategy as {\em top-specific} while the counterpart as {\em bottom-specific}. 
We refer to a model that is trained on diverse domains as a {\em multi-domain model}. 
Based on an extensive evaluation on four representative convolutional neural networks (CNN) architectures, we make two major observations.

\setlist{nolistsep}
\begin{itemize}
    \item Multi-domain models constructed using the bottom-specific strategy could achieve significantly better performance than the ones constructed using the top-specific strategy (i.e., the common practice). Controlled experiments show that this phenomenon can be reproduced with the different number of domains trained together on different backbone architectures using different quantities of domain-specific parameters. 
    
    \item Multi-domain models with few domain-specific parameters from bottom layers can achieve the same, if not better, performance, as independent models trained separately on each domain in terms of the validation accuracy, which introduces the bottom-specific strategy as a strong baseline for model design in MDL. 
\end{itemize}

These observations advocate for people to rethink the design of hard parameter sharing strategies in MDL. Particularly, because top layers of a modern CNN architecture are usually wider, they tend to have higher redundancy and representation capability that is not fully exploited when trained on a single task. 
Several prior studies~\cite{liu2018understanding, yu2020understanding} empirically demonstrate that bottom layers are less redundant than top layers in existing architectures.
These studies raise a potential explanation to our observation: the bottom-specific strategy could achieve better performance than the top-specific strategy because it makes full use of the capacity in top layers while alleviating task interference by increasing the representation power of bottom layers. 
We further validate the explanation via a set of controlled experiments based on network pruning. 
Our experimental evidence and discussions suggest using the bottom-specific strategy as a stronger baseline for model design in MDL.    

\section{Related Work}

\textbf{Multi-Task Learning.} 
Recent works in multi-task learning (MTL) create multi-task models based on popular DNN architectures called {\em backbone architectures}. They fall into either hard parameter sharing or soft parameter sharing \cite{ruder2017overview}. 
Compared with soft parameter sharing where each task still keeps its own model and parameters, hard parameter sharing allows multiple tasks to share some of the model parameters and enjoys the benefits of reduced storage cost and inference latency.
This paper thus focuses on hard parameter sharing.

The most widely-used hard parameter strategy is proposed by Caruana~\cite{caruana1997multitask}, which shares the bottom layers of a model across tasks. For instance, Multi-linear Relationship Networks~\cite{long2017learning} share the first five convolutional layers of AlexNet and use task-specific fully-connected layers for different tasks. Meta Multi-Task Learning~\cite{ruder2019latent} could also be  materialized as a hard parameter sharing structure. Hard parameter sharing may suffer from task interference because tasks compete for the same set of parameters in the shared bottom layers. However, it remains the most popular paradigm due to its effectiveness in reducing the risk of overfitting and storage cost.


\textbf{Multi-Domain Learning.} 
Multi-domain learning (MDL) aims at utilizing a single network to perform target tasks in a diverse set of domains. This paper focuses on MDL and studies how to design a compact model that jointly learns representations for all the domains with a small number of domain-specific parameters. 

There are two types of approaches to developing multi-domain models. The first type of approaches designs various adapter modules (e.g., Batch Normalization Adapter \cite{bilen2017universal} and Residual Adapter \cite{rebuffi2017learning}) and plugs them in the backbone architecture. The entire backbone architecture keeps domain-agnostic and is shared across domains while the adapters are domain-specific. A recent study~\cite{zhao2021and} has shown that the choice of adapters and the locations they are plugged in depend on the set of domains. It leverages neural architecture search to figure out what adapter to use and where to add adapters for a given set of domains.

The second type of approaches follows the common practice of hard parameter sharing in MTL. Some researchers \cite{sarwar2019incremental, he2020multi} proposed to share bottom layers and design sophisticated top layers for each domain. However, it remains an open question whether the effectiveness of sharing parameters in bottom layers in MTL can transfer to MDL. In other words, it is unclear whether the common practice adopted for tasks from a single domain could lead to similarly satisfied performance for tasks from different domains. 
In this work, we aim to answer the open question 
and provide insights on designing a better hard parameter sharing strategy for MDL.

\section{Methodology}\label{sect:method}

Our goal is to study the performance of different hard parameter sharing strategies via controlled experiments, in which the number of tasks, the backbone architectures, the quantity of domain-specific parameters are taken into account.
We first describe three design considerations for the studied sharing strategies in this section and then report experimental settings and results in the following two sections. 

\textbf{Domain-specific parameters in the filter granularity.}
Our experiments focus on image classification tasks, where each task has its own dataset. CNN models naturally become the backbone architectures due to their superior performance on vision tasks. To create multi-domain models, we determine domain-specific parameters in the granularity of filters instead of layers. Specifically, we will compare the performance of multi-domain models created using the common practice (i.e., the top-specific strategy) and its counterparts (i.e., the bottom-specific strategy) given the same targeted amount of domain-specific weights. The filter-level granularity allows us to precisely control the percentage of domain-specific parameters and ensures that each strategy in comparison can have the same percentage in controlled experiments. 
Our experiments show that both the amount of domain-specific parameters and where these parameters come from (e.g., bottom layers or top layers) have a significant impact on the performance of a multi-domain model.   

\textbf{Separate classifier for each domain.}
It is common that different domains expect a different size of outputs or even have diverse prediction goals. 
In our experiments, the backbone architectures use a separate classifier (i.e., the output layer) for each domain to fit the needs of different output dimensions. Although it is still feasible to allow multiple image classification tasks to share the same classifier, we observe serious performance degradation due to the aggressive sharing. Thus, following the practice in prior work~\cite{standley2020tasks}, we adopt a separate classifier for each domain. 

\textbf{Separate batch normalization layers for each domain.}
We use separate Batch Normalization (BN) layers for each domain in our multi-domain models. It is motivated by a prior study~\cite{mudrakarta2018k}, which shows that re-learning a set of scales and biases is sufficient to achieve comparable performance as re-learning the entire set of parameters when a pre-trained model is transferred to another task. 
The scales and biases correspond to parameters in BN layers in typical CNN architectures. In our experiments, we adopt the idea of making BN parameters domain-specific.

Figure~\ref{fig:methodology} illustrates the forward propagation of a multi-domain model given a specific domain. All domains to be learned together share the same backbone architecture.  Each domain has its own BN layers, the output layer that produces logits, and a subset of convolutional filters from the backbone architecture. The remaining convolutional filters are shared by all domains. When a set of filters in a convolutional layer is domain-specific, we use the activation maps produced by these filters to replace the corresponding activation maps from the backbone architecture before the activation maps are fed into the next layer.

\section{Experimental Settings} \label{sect:settings}
To comprehensively and fairly compare the performance of different hard parameter sharing strategies, we need to consider several potentially influential factors including the backbone architectures, the set of domains and domain-specific parameters, parameter initialization, and hyper-parameters. We next explain each factor in detail.

\textbf{Backbone architectures.}
Our experiments use four popular CNNs, MobileNetV2, ResNet50, MNasNet, and SqueezeNet. When creating multi-domain models based on each backbone architecture, we add a separate set of BN layers and a separate output layer for each domain. We also select a subset of filters as domain-specific parameters and share the rest among all domains to be learned jointly. 

\textbf{Datasets.}
We conduct extensive experiments on five fine-grained image classification tasks. For simplicity, we call it the \textit{FGC benchmark}. 
We also verify our observations on Decathlon \cite{rebuffi2017learning}, which contains ten well-known datasets from diverse visual domains. 

\textbf{Domain-specific parameters.}
We experiment with different quantities of domain-specific parameters and sharing strategies. Specifically, we pick different quantities of domain-specific filters such that the number of weights in these filters accounts for 0\% to 100\% of the total number of convolution parameters in the backbone architecture. We use a step size of 10\%. Given the same percentage of domain-specific parameters, we compare three hard parameter sharing strategies: 
\begin{itemize}
\item {\em top-specific.} This strategy shares filters in bottom layers and makes filters in top layers domain-specific. It is the common practice used in hard parameter sharing~\cite{ruder2017overview, standley2020tasks}. 
\item {\em bottom-specific.} It shares filters in top layers while making filters in bottom layers domain-specific. This is a direct counterpart of the top-specific strategy. 
\item {\em random.} It randomly selects a subset of filters from all convolutional layers as domain-specific parameters and shares the rest. 
\end{itemize}

\textbf{Initialization.}
Our goal is to eliminate randomness caused by initialization during controlled experiments to ensure the fairness of the comparison and the reproducibility of the experimental results. 
All multi-domain models are initialized with their corresponding backbone weights pre-trained on ImageNet, while the domain-specific classifiers are pre-trained on each domain separately used to initialize the classifiers in multi-domain models. 


\section{Results and Analysis}\label{sect:results}
\subsection{Comparisons between sharing strategies on FGC} \label{sect:firstdiscovery}
Our first surprising discovery is that when constructing multi-domain models, using separate bottom-layer parameters could achieve much better performance than using separate top-layer parameters, which contradicts the common belief in hard parameter sharing in MDL.

\begin{table*}[htb]
\footnotesize 
\begin{center}
\caption{The validation accuracy of the 12 multi-domain models (4 backbone architectures $\times$ 3 sharing strategies) on FGC with the same amount of domain-specific parameters for each domain (20\% of the total number of convolution parameters in the backbone model). ``independent'' reports the results of independent models trained on each domain separately. } 
\label{table:frontbetter}
\vspace{5pt}
\begin{tabular}{|c|c|c|p{1.2cm}<{\centering}p{1.2cm}<{\centering}p{1.2cm}<{\centering}p{1.2cm}<{\centering}p{2cm}<{\centering}|}
\hline
Architectures   & \# Params (M)             & Sharing Strategy & Aircraft & Birds & Cars & Dogs & Indoor Scenes \\ \hline\hline
\multirow{4}{*}{MobileNetV2} & 10.67 & top-specific             &   0.8536      &      0.6714  &    0.8145      &    0.6006      &   0.5985          \\
                             & 10.61 & random          &   0.8617      &      0.6699  &    0.8259      &    0.6021      &   0.6013  \\
                             & 10.63 & bottom-specific          &\textbf{0.8782}&\textbf{0.6920}&\textbf{0.8506}&\textbf{0.6186} &  \textbf{0.6119}   \\ \cline{2-8}
                             & 17.52 & independent     &   0.8749      &      0.6920  &    0.8496      &    0.6202      &   0.6074  \\ \hline\hline
\multirow{4}{*}{ResNet50}    & 52.84 & top-specific             &   0.8464      &      0.6243  &    0.7887      &    0.5625      &  0.5530           \\
                             & 52.74 & random          &   0.8650      &      0.6740  &    0.8218      &\textbf{0.6211} &  0.6037           \\ 
                             & 52.97 & bottom-specific          &\textbf{0.8657}&\textbf{0.6925}&\textbf{0.8363}&    0.6128      &  \textbf{0.6067}  \\ \cline{2-8}
                             & 127.78 & independent     &   0.8680      &      0.6799  &    0.8419      &    0.6248      &   0.6037  \\\hline\hline
\multirow{4}{*}{MNasNet}     & 12.31 & top-specific             &   0.8422      &      0.6369  &    0.7747      &    0.5880      &  0.6104           \\
                             & 12.34 & random          &   0.8482      &      0.6634  &    0.7946      &    0.6080      &  0.6082           \\ 
                             & 12.24 & bottom-specific          &\textbf{0.8596}&\textbf{0.6740}&\textbf{0.8081}&\textbf{0.6212} &  \textbf{0.6164}  \\ \cline{2-8}
                             & 21.91 & independent     &   0.8668      &      0.6660  &    0.8174      &    0.6177      &   0.6037  \\\hline\hline
\multirow{4}{*}{SqueezeNet}  & 3.91 & top-specific             &   0.7555      &      0.5666  &    0.6423      &    0.5034      &  0.5000           \\
                             & 3.90 & random          &   0.7702      &\textbf{0.5817}&   0.6743      & \textbf{0.5293}&  0.5119           \\
                             & 3.90 & bottom-specific         &\textbf{0.7705}&      0.5733  &\textbf{0.6764} &    0.5218      &  \textbf{0.5131}  \\ \cline{2-8}
                             & 6.24 & independent     &   0.7819      &      0.6018  &    0.7016      &    0.5427      &   0.5321  \\ \hline
\end{tabular}
\end{center}
\vspace{-10pt}
\end{table*}

\begin{figure}[htb]
\centering
\subfigure[MobileNetV2]{
\includegraphics[width=0.34\textwidth]{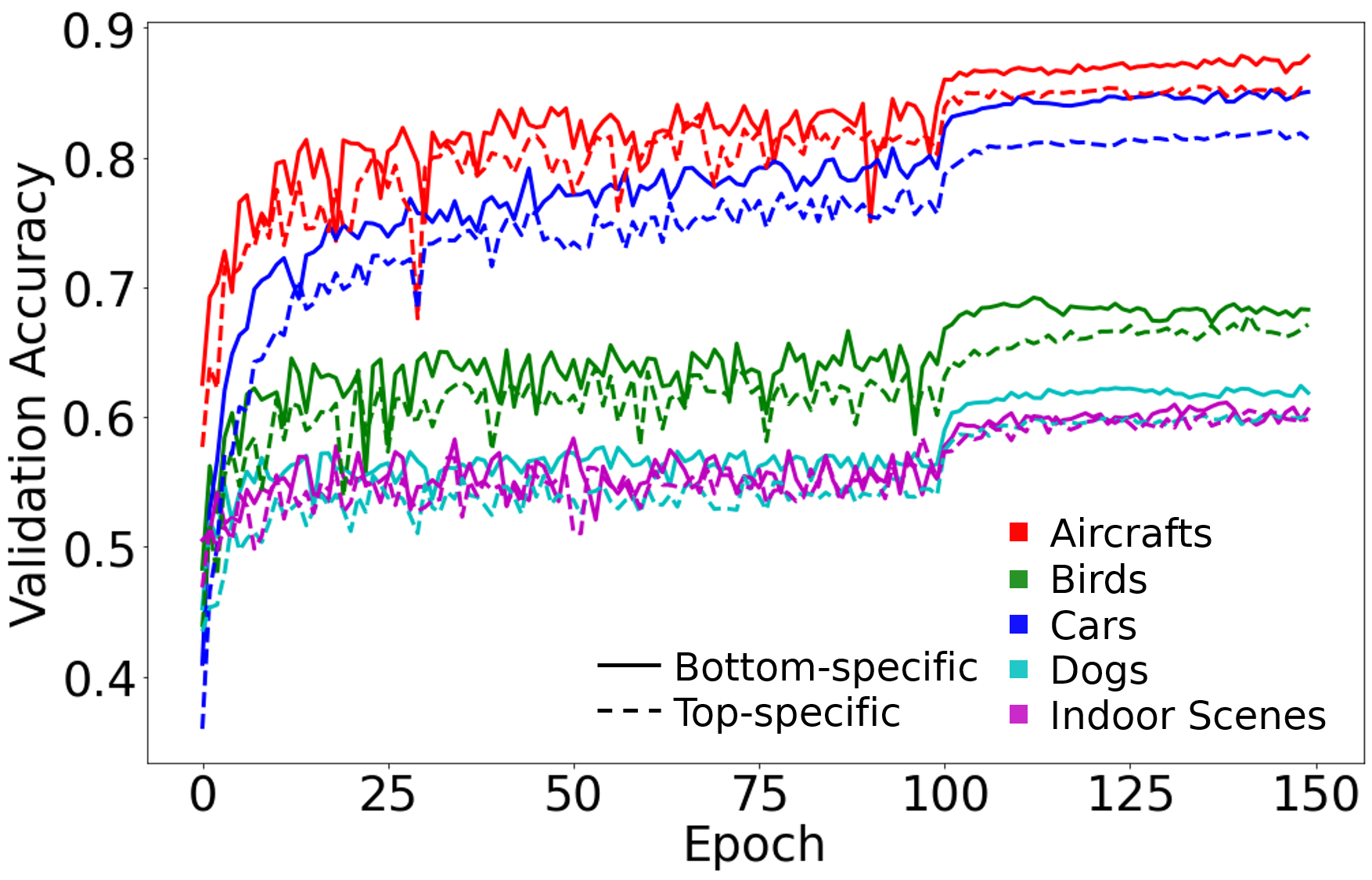}
}
\subfigure[ResNet50]{
\includegraphics[width=0.34\textwidth]{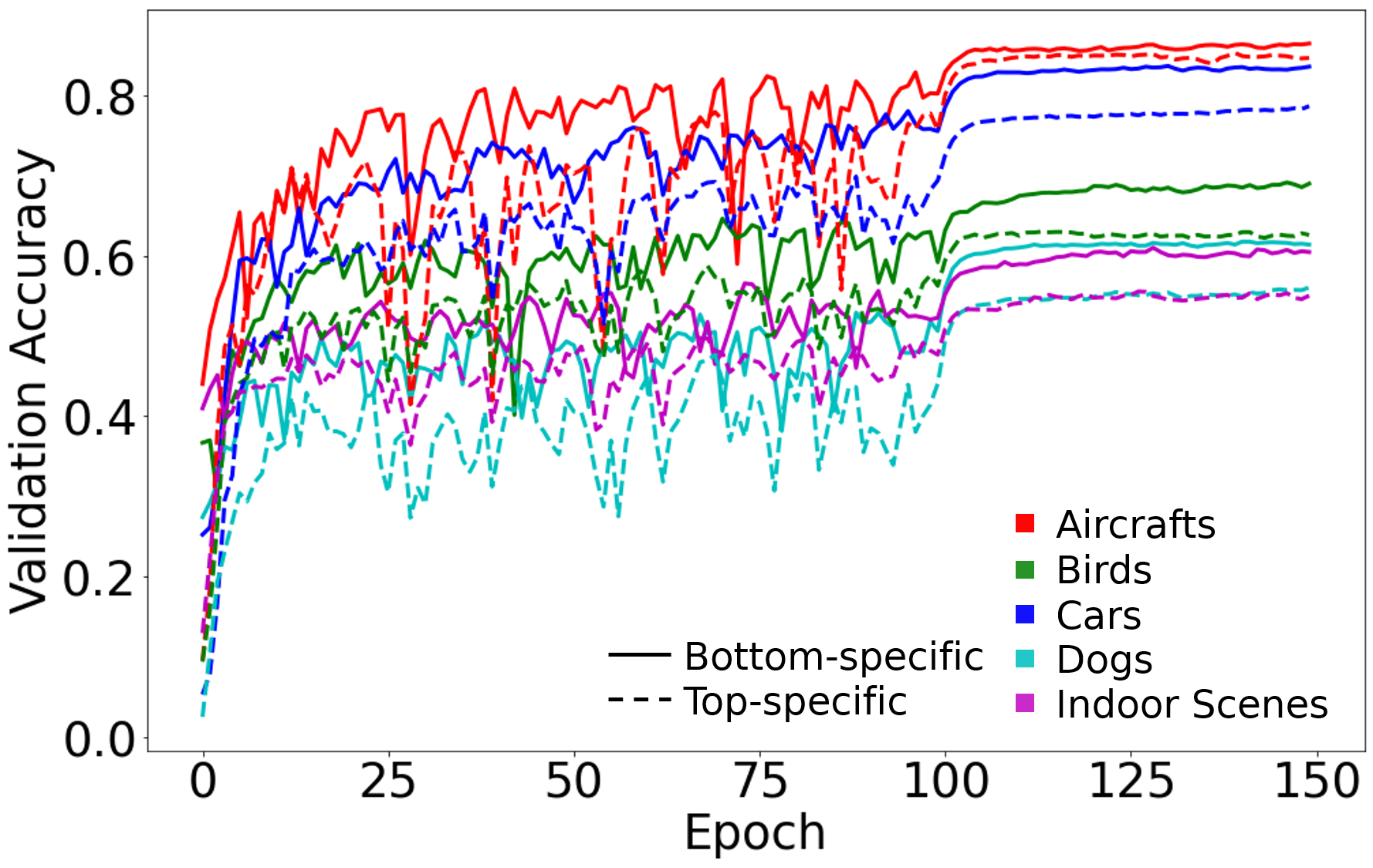}
}
\vspace{-0.1in}
\caption{The accuracy curves of multi-domain models created with the bottom-specific or top-specific strategy on the five domains in FGC (present by five colors). The percentage of domain-specific parameters for each domain is 20\% of the total amount of convolution parameters in the backbone model, (a) MobileNetV2 and (b) ResNet50.
}
\label{fig:frontbetter}
\end{figure}

In the experiment, we construct three multi-domain models from each backbone architecture using the three sharing strategies (top-specific, bottom-specific, and random) and train these models on FGC. 
We strictly control the percentage of domain-specific parameters to be the same for the multi-domain models created using the three strategies.

Table \ref{table:frontbetter} reports the validation accuracy of the 12 multi-domain models (4 backbone architectures $\times$ 3 sharing strategies) on the five domains. 
The number of domain-specific parameters is 20\% of the total amount of convolution parameters in the backbone model. The rows ``random'' report the mean accuracy of two multi-domain models constructed using the random strategy.
The results indicate that the bottom-specific strategy consistently achieves better performance than the top-specific one and outperforms the random strategy in most cases.

\textbf{Different backbone architectures.} 
We observed the same phenomenon using different backbone architectures. Figure~\ref{fig:frontbetter} shows the accuracy curves of multi-domain models created with different strategies. The accuracy curves of the bottom-specific strategy are always above the ones with the top-specific strategy on all five domains with different backbone architectures, indicating the better performance of the bottom-specific strategy.
The curves of multi-domain models built on MNasNet and SqueezeNet show a similar pattern.

\textbf{Different number of domains.} 
The same observation also holds with the different number of domains. Table~\ref{table:frontbetter} shows the performance when all five domains are trained together and Table~\ref{table:tasksnumber} presents the results when fewer domains are used to jointly train multi-domain models built with MobileNetV2. 
The results suggest that the bottom-specific strategy consistently yields a better prediction performance than the top-specific strategy no matter the number of domains to learn jointly.

\begin{table}[h]
\footnotesize 
\begin{center}
\caption{The validation accuracy of multi-domain models trained on different number of domains. The amount of domain-specific parameters for each domain accounts for 20\% of the total number of convolution parameters in the backbone architecture MobileNetV2.}\label{table:tasksnumber}
\vspace{5pt}
\begin{tabular}{|c|cccc|}
\hline
Strategy & Aircraft & Birds & Cars & Dogs \\ \hline\hline
top-specific             &   0.8719      &    -  &    0.8390      &    -        \\
random          &   0.8762      &    -  &    0.8387      &    -        \\
bottom-specific         &\textbf{0.8767}&    -  &   \textbf{0.8460}   &    -   \\ \hline
top-specific             &   0.8548      &      0.6655  &    0.8336      &   -        \\
random          &\textbf{0.8687}&      0.6716  &    0.8357      &   -        \\ 
bottom-specific          &   0.8677      &\textbf{0.6786}&\textbf{0.8474}&    -   \\ \hline
top-specific             &   0.8629      &      0.6723  &    0.8256      &    0.5948        \\
random          &   0.8683      &      0.6902  &    0.8318      &    0.6118        \\ 
bottom-specific          &\textbf{0.8713}&\textbf{0.6917}&\textbf{0.8423}&\textbf{0.6186}   \\ 
\hline
\end{tabular}
\end{center}
\vspace{-5pt}
\end{table}

\begin{figure*}[h]
\centering
\subfigure[FGVC Aircraft]{
\includegraphics[width=0.165\linewidth]{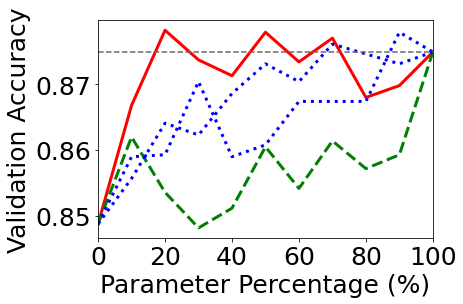}
}
\subfigure[CUB-200-2011]{
\includegraphics[width=0.165\linewidth]{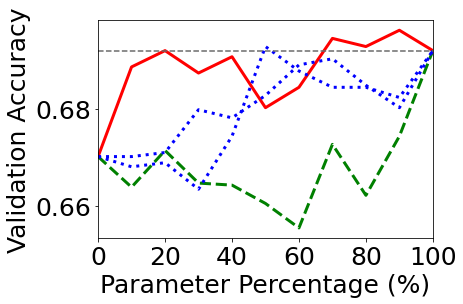}
}
\subfigure[Stanford Cars]{
\includegraphics[width=0.165\linewidth]{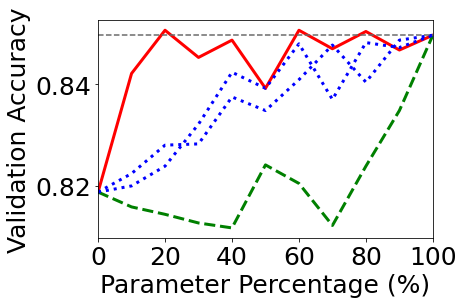}
}
\subfigure[Stanford Dogs]{
\includegraphics[width=0.165\linewidth]{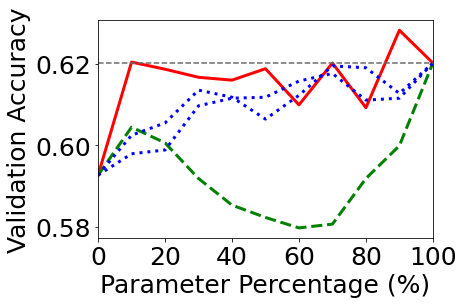}
}
\subfigure[MIT Indoor Scenes]{
\includegraphics[width=0.25\linewidth]{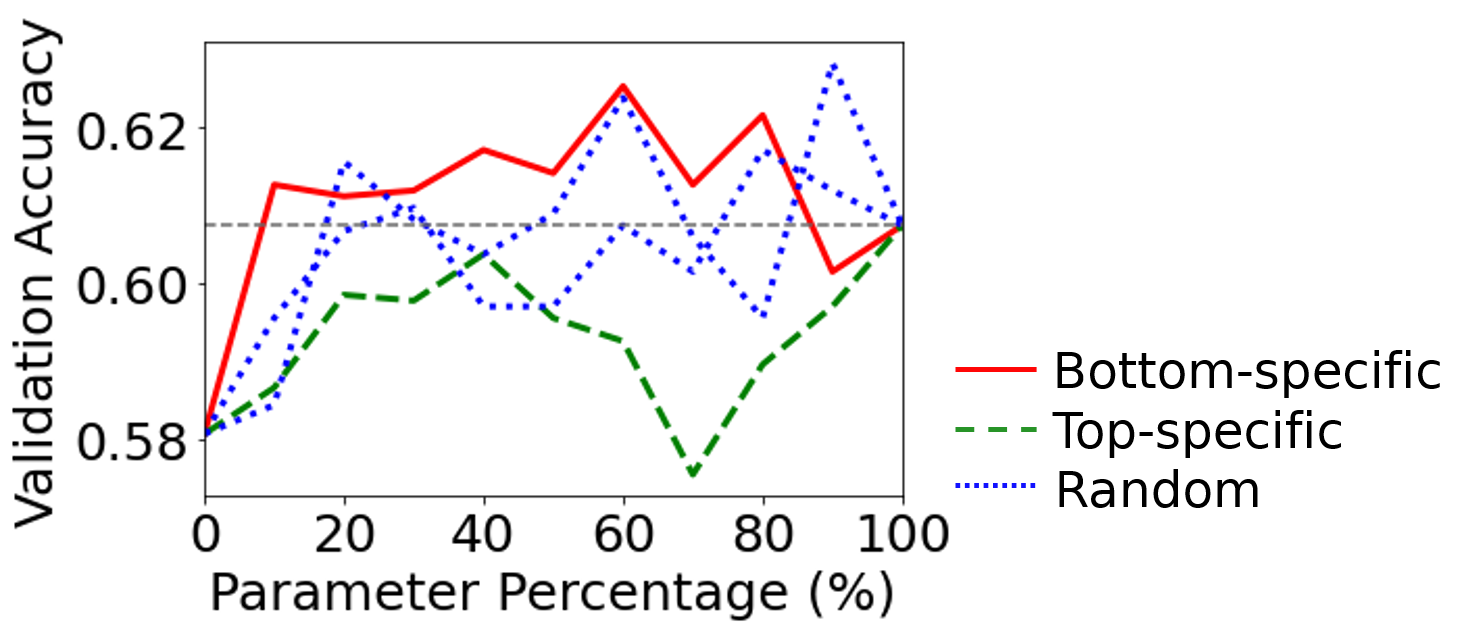}
}
\vspace{-0.1in}
\caption{The validation accuracy trends with an increasing percentage of domain-specific parameters (0\% to 100\% of the total convolution parameters in MobileNetV2) for the five domains in FGC. In each line chart, the four lines correspond to applying different hard parameter sharing strategies.}
\label{fig:paramquantity}
\vspace{-0.1in}
\end{figure*}

\textbf{Different number of domain-specific parameters.} 
The superiority of the bottom-specific strategy still holds with different quantities of domain-specific parameters. Figure \ref{fig:paramquantity} shows the validation accuracy of the multi-domain models whose domain-specific parameters account for 0\% to 100\% of the total amount of convolution parameters of their backbone architecture MobileNetV2. Note that picking 0\% domain-specific parameters results in a multi-domain model with only separate BN layers and domain-specific classifiers, while possessing 100\% domain-specific parameters is equivalent to building up a completely independent model for each domain. The results show that the performance of the bottom-specific strategy is always higher than the top-specific one, as indicated by the significant gap between the red and green curves. Besides, randomly selecting domain-specific filters also consistently produces higher validation accuracy than the top-specific strategy but is worse than the bottom-specific one in most cases. 

\subsection{Comparisons between sharing strategies on Decathlon} \label{sect:decathlon}
The same phenomenon can be observed when constructing multi-domain models on Decathlon with the bottom-specific and the top-specific sharing strategy. 

\begin{figure}[htb]
\centering
\includegraphics[width=0.22\textwidth]{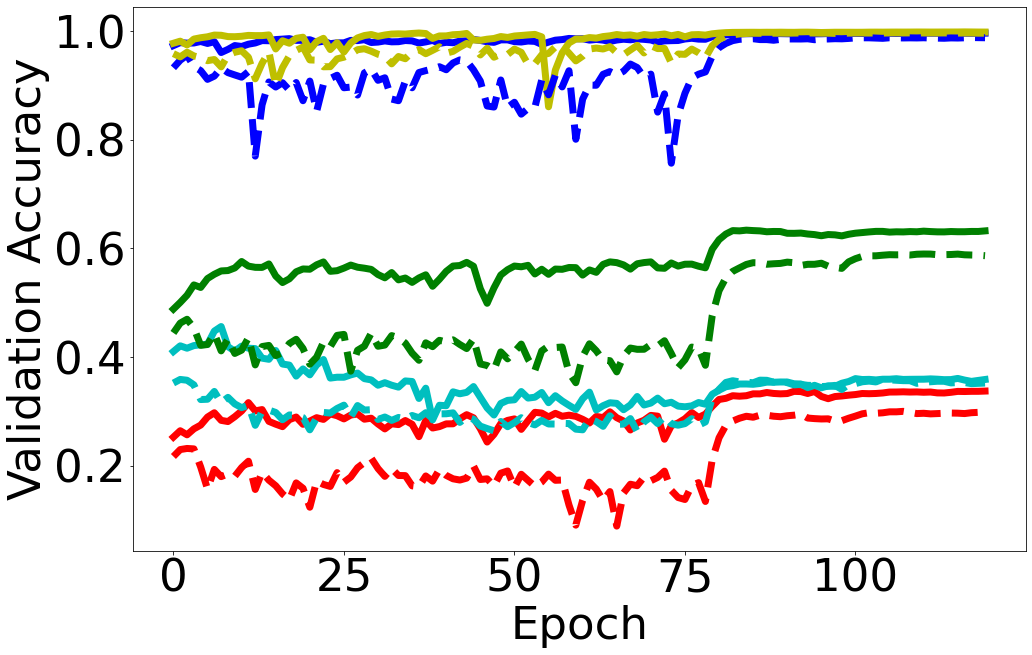}
\includegraphics[width=0.22\textwidth]{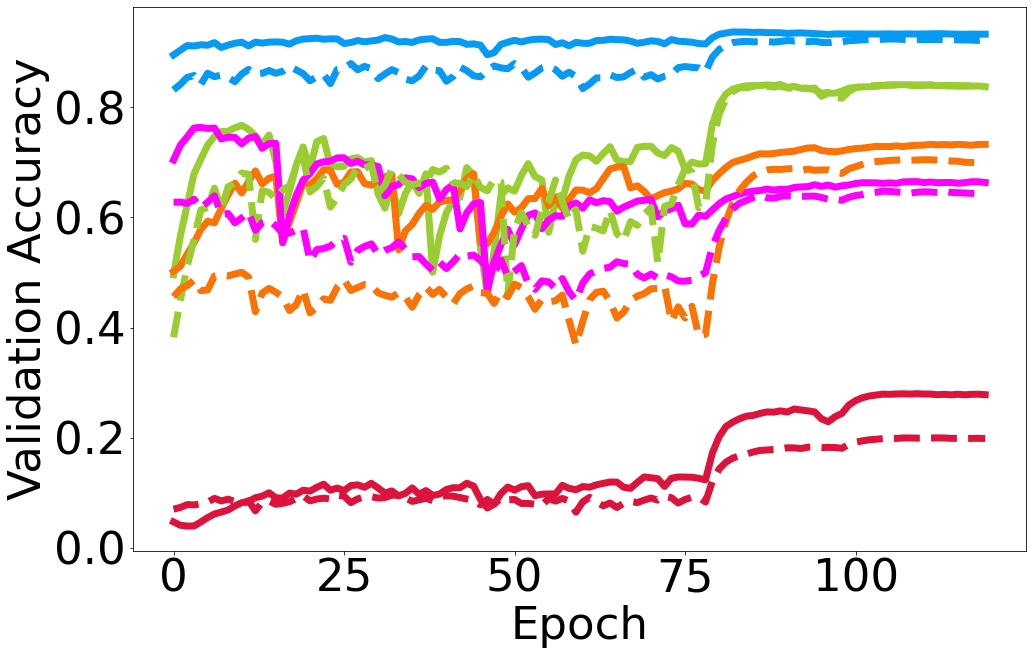}
\includegraphics[scale=0.25]{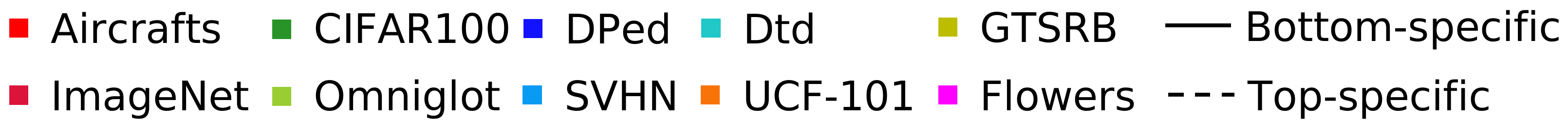}
\caption{The accuracy curves of multi-domain models created with the bottom-specific or top-specific strategy on the ten domains in Decathlon (present by ten colors). The percentage of domain-specific parameters for each domain is 20\% of the total amount of convolution parameters in the backbone model MobileNetV2. 
}
\label{fig:decathlon}
\vspace{-0.1in}
\end{figure}

Figure~\ref{fig:decathlon} shows the accuracy curves of multi-domain models built on MobileNetV2 with different sharing strategies under the same amount of domain-specific parameters (20\% of the total number of convolution parameters in the backbone model). It can be seen that the bottom-specific strategy can produce better accuracy performance than the top-specific one for all or most of the ten domains, which is consistent with the observations on FGC. 

\subsection{Comparison with independent models} \label{sect:seconddiscovery}
Our second discovery is that multi-domain models with a relatively small proportion of parameters selected from bottom layers for each domain can achieve competitive performance with independent models trained on each domain separately. 

In Figure~\ref{fig:paramquantity}, the performance of independent models correspond to the point where the percentage of domain-specific parameters is 100\%. 
Overall, multi-domain models constructed with the bottom-specific strategy can achieve competitive validation accuracy as independent models when the percentage of domain-specific parameters is over 20\% for all the five domains. 
This observation is consistent with the well-recognized benefits of MDL in reducing overfitting and improving prediction accuracy.

\section{Discussions}
\label{sect:discussions}

We summarize the two main observations from our experiments as follows.
\setlist{nolistsep}
\begin{itemize}
    \item  Multi-domain models with domain-specific parameters from bottom layers could achieve better performance than the ones with the same amount of domain-specific parameters from top layers.
    \item  Multi-domain models with a relatively small quantity of domain-specific parameters from bottom layers achieve the same, if not better, performance as their independent counterparts.
    
\end{itemize}

\textbf{Why the bottom-specific sharing strategy outperforms the top-specific one?} A potential explanation is that top layers of a modern CNN architecture are usually much wider than bottom layers and thus have a higher representation capability. Prior studies~\cite{liu2018understanding, yu2020understanding} have shown that bottom layers have less redundancy than top layers in existing architectures. When tackling multiple domains together, top layers may have sufficient capacity to learn diverse features while the bottom layers are easily distracted by different domains during training. The bottom-specific strategy could achieve better performance than the top-specific strategy because it makes full use of the capacity in top layers while alleviating task interference by increasing the representation power of bottom layers. 

\textbf{Why owning a small proportion of domain-specific parameters from bottom layers is sufficient to surpass independent models?}
The reason comes from two aspects. Firstly, DNNs are well-known to be over-parameterized~\cite{frankle2018lottery, allen2019convergence}. It is reasonable to assume that a single DNN, especially its top layers, can largely accommodate the representation requirements of multiple domains by taking advantage of the redundant parameters that are not fully exploited on a single domain. A small portion of domain-specific parameters from bottom layers increases the representation power of bottom layers, alleviates domain interference, and thus improves model performance. Secondly, the commonalities among tasks serve as implicit data augmentation ~\cite{ruder2017overview} and avoid overfitting. This is consistent with the common belief about the benefits of MDL.


\section{Conclusions} 
In this work, we revisit the common practice in hard parameter sharing for multi-domain learning (MDL) and conduct an empirical study on datasets from different visual domains to compare the performance of different sharing strategies. Experiments show that the common sharing strategy is outperformed by its direct counterpart--that is, selecting domain-specific parameters from bottom layers rather than top layers. The counterpart can also achieve competitive performance compared with independent models. We further provide explanations for the observations.

\bibliographystyle{IEEEbib}
\small
\bibliography{icme2022template}

\end{document}